\documentclass[10pt]{article} 
\usepackage[preprint]{tmlr}

\usepackage{amsmath,amsfonts,bm}

\def\eqref#1{equation~\ref{#1}}

\def\1{\bm{1}}

\DeclareMathAlphabet{\mathsfit}{\encodingdefault}{\sfdefault}{m}{sl}
\SetMathAlphabet{\mathsfit}{bold}{\encodingdefault}{\sfdefault}{bx}{n}

\usepackage{hyperref}
\usepackage{url}
\usepackage[utf8]{inputenc}          
\usepackage[T1]{fontenc}             
\usepackage{lmodern}                 
\usepackage[whole]{bxcjkjatype}      
\usepackage{placeins}                
\usepackage{flafter}                 
\usepackage{authblk}
\usepackage{amsmath}                 
\usepackage{amssymb}                 
\usepackage{amsfonts}
\usepackage{booktabs}                
\usepackage{multirow}
\usepackage{graphicx}
\usepackage{setspace}
\usepackage{microtype}

\include{dataset}
\include{brier}
\include{brier_difference}
\include{ece}
\include{profit}
\include{confidence}
\newcommand{\var}[1]{\csname #1\endcsname}

\title{Foresight Learning for SEC Risk Prediction}

\author{\name Benjamin Turtel \footnote{Contact: ben@lightningrod.ai}, Paul Wilczewski, Danny Franklin, Kris Skotheim \\ \centering \addr \href{https://lightningrod.ai}{Lightning Rod Labs} }

\begin{document}

\maketitle

\begin{abstract}
Risk disclosures in SEC filings describe potential adverse events but rarely quantify their likelihood, limiting their usefulness for probabilistic analysis. A central obstacle is the absence of large-scale, risk-level supervision linking disclosed risks to realized outcomes.
LightningRod AI provides a fully automated data generation pipeline that converts qualitative SEC risk disclosures into temporally grounded supervision using only public data. For each filing, the pipeline generates firm-specific, time-bounded risk queries from the Risk Factors section and labels them by resolving outcomes against subsequent disclosures. The same platform supports applying this pipeline to arbitrary document collections, enabling scalable outcome-linked supervision without human annotation.
Using this dataset of risk queries and outcomes grounded in SEC filings, we train a compact large language model to estimate the probability that a disclosed risk will materialize within a specified horizon. Despite its modest size, the resulting model substantially improves over pretrained and heuristic baselines, and outperforms frontier general-purpose models, including GPT-5, on probabilistic accuracy and calibration.
More broadly, this work demonstrates that automated future-based supervision can fine-tune language models to frontier-level performance using only raw, chronological, in-domain text - without proprietary data, external corpora, or manual annotation. Using SEC filings as a case study, we show that foresight learning enables the training of compact, domain-specific expert models that outperform frontier general-purpose models while remaining deployable on a single GPU. This result suggests a general pathway for learning calibrated, decision-relevant signals from naturally occurring enterprise documents. To support transparency and reproducibility, we open-source the evaluation dataset used in this study.
\footnote{Dataset available at: \href{}{https://huggingface.co/datasets/LightningRodLabs/sec\_risk\_questions\_test\_set}}
\end{abstract}

\section{Introduction}

\label{sec:intro}Public companies are required to disclose a wide range of risks in periodic filings with the U.S. Securities and Exchange Commission (SEC). These disclosures describe potential adverse events, operational challenges, regulatory constraints, and strategic uncertainties, but they are almost entirely qualitative: firms explain what could happen, not how likely it is to occur. As a result, investors and risk practitioners must reason about risk likelihoods from narrative text without systematic or calibrated guidance.

We study the problem of risk materialization prediction from SEC disclosures. Given a specific disclosed risk, the task is to estimate the probability that the risk will materialize within a future reporting horizon. Framing prediction at the level of individual risks enables fine-grained probabilistic reasoning grounded directly in firms’ own disclosures, rather than document-level proxies or market reactions.

A fundamental challenge is the lack of labeled data. Risk outcomes are rarely annotated ex ante, and the space of possible risks is unstructured, heterogeneous, and firm-specific. Manual labeling at scale is infeasible. To address this, we introduce a fully automated data generation pipeline built on LightningRod AI that uses only public SEC filings. The pipeline generates firm-specific, time-bounded risk queries and resolves their outcomes by linking to subsequent disclosures, producing large-scale risk-level supervision without human annotation.

We train a language model using Foresight Learning as formulated as Future-as-Label \cite{turtel2026}, training it to output calibrated probabilities aligned with realized future outcomes. While Foresight Learning has previously been applied to forecasting events in news, we show that it extends to long, unstructured regulatory disclosures describing abstract and heterogeneous risks.

Empirically, the resulting model substantially outperforms naive heuristics and strong pretrained baselines across probabilistic accuracy and calibration. These results demonstrate that learning directly from future disclosures allows language models to extract meaningful, well-calibrated risk signals without human annotation or proprietary data.

Our contributions are fourfold:

\begin{enumerate}
\item We formulate risk materialization prediction as a probabilistic forecasting task grounded in corporate disclosures;
\item We introduce a fully automated pipeline for generating large-scale, risk-level supervision from public SEC filings;
\item We show that Foresight Learning enables language models to learn nontrivial probabilistic signals about future risk realization from unstructured regulatory text; and
\item We demonstrate a general, scalable approach for training domain-specific expert models from unstructured chronological data using future-resolved supervision, without manual annotation or proprietary data.
\end{enumerate}

\section{Related Work}

A growing literature studies the informational content of risk disclosures in SEC filings and related corporate reports. Prior work shows that textual risk disclosures improve predictions of outcomes such as financial distress, returns, volatility, and firm performance when incorporated into machine learning models at the document level \cite{zavitsanos2025} \cite{hajekmunk2025}. These results establish that narrative risk disclosures contain forward-looking signal, but typically treat risk text as an aggregate input rather than modeling individual disclosed risks.

More targeted evidence comes from studies of specific risk domains, such as cybersecurity, which find that firms’ mandatory cyber risk disclosures are predictive of subsequent realized incidents \cite{kesari2022}. While this work demonstrates that disclosed risks can foreshadow concrete adverse events, it focuses on narrow risk categories and does not address probabilistic estimation across heterogeneous risk types.

Recent advances in domain-specific large language models for finance show strong performance on a wide range of financial tasks, highlighting the ability of LLMs to capture nuanced semantics in complex regulatory text \cite{caillaut2025}. However, most applications frame prediction as classification or regression on contemporaneous labels rather than as calibrated probability estimation of future events.

In contrast, we formulate risk materialization as a conditional probability estimation problem at the level of individual disclosed risks. We construct supervision by resolving risk outcomes using future SEC filings and train an LLM using Foresight Learning to produce calibrated probabilities of future risk realization. This shifts the focus from aggregate document-level prediction to fine-grained, risk-level uncertainty modeling grounded directly in firms’ forward-looking disclosures.

Our approach builds on Foresight Learning, introduced in prior work \cite{turtel2026}, which studies how models can be trained using supervision derived from the eventual resolution of events rather than contemporaneous labels. Related outcome-based approaches, such as reinforcement learning with verifiable rewards (RLVR), similarly rely on supervision from realized outcomes but have largely been explored in closed-world domains such as programming and mathematics, where outcomes are unambiguous. We extend Foresight Learning to corporate risk disclosures, showing that it scales to long, unstructured regulatory text and supports calibrated probability estimation over abstract and heterogeneous risk events.

\section{Data and Problem Setup}

\subsection{Risk Queries}

Each data point in our dataset is a \emph{risk query}: a firm-specific, time-bounded, and falsifiable natural-language proposition describing the potential materialization of a disclosed risk. Risk queries are derived from the \emph{Risk Factors} section of an SEC filing and correspond to legal, financial, operational, or strategic risks over a predefined future horizon.

The following are exact examples of queries from the evaluation data:

\begin{itemize}
\item Will DraftKings Inc. (NASDAQ: DKNG) disclose in a SEC filing that it has ceased all sports betting or iGaming operations in any U.S. state where it was active as of November 8, 2024, due to a state-level legislative or regulatory prohibition by June 30, 2025?
\item Will Ceva, Inc. (NASDAQ: CEVA) announce a new share repurchase program or an expansion of an existing share repurchase program between November 8, 2024, and June 30, 2025?
\end{itemize}

All queries in the dataset are generated automatically without human editing or filtering.

\subsection{SEC Filings and Risk Query Construction}

Our primary data source consists of periodic filings submitted to the U.S. Securities and Exchange Commission (SEC), with a focus on Forms 10-K and 10-Q, which provide standardized, legally mandated disclosures of firm-specific risks at regular intervals. We select filings from Q1 2024 and Q4 2024 as source documents in order to construct a temporally well-defined prediction task with sufficient forward coverage for outcome resolution. This design ensures that all risk queries are generated from information available at a clearly specified disclosure date and can be evaluated against subsequent filings within a fixed future horizon.

Risk queries are generated using a retrieval-augmented generation (RAG) pipeline built on Gemini. For each firm--period, the pipeline retrieves content from the \emph{Risk Factors} section of the source filing and uses Gemini-2.5-Flash to generate natural-language queries phrased as concrete, falsifiable propositions. Queries may correspond to explicitly stated risks or to risks that are clearly implied by the disclosure text, and each query specifies both an event definition and a future prediction horizon.

The resulting prediction task is to estimate the probability that the queried risk will materialize within the specified horizon, conditional solely on information available at the time of the source filing.

\subsection{Risk Materialization and Labeling}

We define risk materialization as the explicit disclosure in subsequent SEC filings that a previously stated risk has occurred or progressed in a manner consistent with the original query. Materialization does not require the outcome to be extreme or financially catastrophic; rather, it requires that the firm acknowledges the occurrence or advancement of the underlying risk event.

Observable outcomes are determined exclusively from future SEC filings for the same firm. These disclosures may include financial outcomes (e.g., changes in revenues, costs, or margins), operational events (e.g., supply chain disruptions or facility closures), strategic actions (e.g., acquisitions or divestitures), legal or regulatory developments, and governance or personnel changes. All labels are grounded strictly in firms’ own disclosures rather than external data sources or market reactions.

To generate supervision at scale, we label each risk query using Gemini-2.5-Flash with retrieval-augmented generation over future SEC filings. For a given query, we retrieve filings that (i) are strictly dated after the source filing and (ii) fall within the specified prediction horizon. Based on the retrieved passages, the model determines whether the queried event is disclosed as having occurred.

Labels are binary: a value of 1 indicates that the risk materialized within the horizon, while 0 indicates that it did not. Although this automated labeling process introduces noise, it enables scalable supervision and reflects the inherent ambiguity and subjectivity of corporate risk disclosures. Because labels are derived from future disclosures rather than contemporaneous annotations, the task naturally favors probabilistic prediction and calibrated uncertainty over hard classification.

\subsection{Dataset Statistics and Splits}

The final dataset consists of 6{,}109 risk queries derived from 2{,}820 unique SEC filings and covering 1{,}953 distinct publicly traded firms. Each query is associated with a binary materialization label.

We split the data into training and test sets based on filing dates to preserve temporal ordering. The training set contains 5{,}609 samples (67.1\% non-materialization, 32.9\% materialization) derived from filings dated between January 3, 2024 and November 7, 2024. The test set contains 500 samples (66.2\% non-materialization, 33.8\% materialization) derived from filings dated between November 7, 2024 and December 27, 2024.

Across the full dataset, filing dates span 359 days. This temporally ordered split ensures that models are evaluated on genuinely future risk disclosures relative to the training data, closely mirroring real-world deployment settings.

\subsection{Risk Category Distribution}

For analysis purposes, each risk query is assigned a high-level risk category using an LLM-based classifier. Categories capture broad classes of disclosed risks rather than fine-grained taxonomies.

Table~\ref{tab:risk-category-distribution} summarizes the distribution of risk queries by category.

\begin{table}[htbp]
\centering
\begin{tabular}{lrr}
\hline
Risk Type & Count & Percent \\
\hline
Expenses, Costs \& Balance Sheet Metrics & 1521 & 24.90\% \\
Revenue \& Sales Performance & 1050 & 17.19\% \\
Strategic Actions \& Corporate Transactions & 776 & 12.70\% \\
Profitability \& Margins & 706 & 11.56\% \\
Governance, Compliance \& Internal Controls & 688 & 11.26\% \\
Legal \& Regulatory Events & 641 & 10.49\% \\
Regulatory Approvals \& Product Development & 331 & 5.42\% \\
Operational Events \& Disruptions & 285 & 4.67\% \\
Other & 111 & 1.82\% \\
\hline
\end{tabular}
\caption{Distribution of risk queries by high-level risk category.}
\label{tab:risk-category-distribution}
\end{table}

These categories are used only for descriptive and diagnostic analysis and do not constrain model inputs or outputs.

\section{Model and Training}

\subsection{Learning Framework}

We frame risk materialization prediction as a probabilistic forecasting task in which supervision is provided by the eventual resolution of events rather than contemporaneous labels. For each risk query, the model observes only information available up to the source filing date $t$ and predicts whether the queried event will materialize by a later time $s > t$.

We adopt the Foresight Learning framework of \cite{turtel2026}, which formalizes learning from temporal data under causal information constraints. In this framework, predictions are made under a masked information state that excludes post-$t$ information, while outcome resolution is performed using future disclosures unavailable at prediction time. This preserves the temporal asymmetry of forecasting even when training is performed offline on resolved events.

Within this framework, the model is trained to produce calibrated probabilistic forecasts, with learning driven by outcome-based supervision derived from future disclosures rather than fixed labeled examples.

\subsection{Base Model}

Our base model is Qwen3-32B, a 32-billion-parameter decoder-only transformer pretrained on a mixture of general-domain and technical text. We select this model for its strong performance on long-context reasoning and its suitability for further adaptation via reinforcement-style fine-tuning.

\subsection{Model Inputs}

For each example, the model input consists of two components:
\begin{enumerate}
    \item a concise natural-language summary of the \emph{Risk Factors} section from the source SEC filing, and
    \item a risk query describing a specific, well-defined future risk event and prediction horizon.
\end{enumerate}

These components are concatenated using a fixed prompt template. No structured financial data, historical prices, or external context is provided. Rather than supplying the full \emph{Risk Factors} section verbatim, we use summary-level representations generated during data construction. This choice ensures consistency across filings of varying length while retaining salient risk information. All inputs fit within the model’s context window, and no truncation or sliding-window techniques are required.

\subsection{Training Objective and Optimization}

The model outputs a single scalar probability in $[0,1]$, interpreted as the likelihood that the queried risk will materialize within the specified horizon. Training maximizes the expected negative Brier score under the model’s predicted probabilities. This objective rewards accurate and well-calibrated probabilistic forecasts while penalizing overconfident incorrect predictions.

We optimize this objective using Generalized Reinforcement Policy Optimization (GRPO), which updates the model to minimize expected scoring loss while maintaining stability relative to the pretrained initialization. Unless otherwise noted, results reported in Section~4 correspond to the final checkpoint at step~144, which we refer to as the \emph{trained model}.

\section{Results}

We evaluate model performance on held-out risk queries using standard metrics for probabilistic prediction: Brier score, Brier skill score (BSS), and expected calibration error (ECE). Brier skill score is computed relative to the naive baseline and measures the fractional improvement in Brier score over this reference; positive values indicate better performance, while negative values indicate degradation. Lower Brier score and ECE indicate better probabilistic accuracy and calibration, respectively.

The naive baseline predicts a constant probability equal to the empirical base rate of risk materialization in the training data. Each model is evaluated using the same input context and prompt template. The held-out evaluation set used for all reported results is available publicly to support reproducibility; training data and model parameters are not released.

\subsection{Aggregate Probabilistic Performance}

Table~\ref{tab:aggregate-performance} summarizes overall performance on the test set.

\begin{table}[htbp]
\centering
\begin{tabular}{lccc}
\hline
Model & Brier~$\downarrow$ & Brier Skill Score~$\uparrow$ & ECE~$\downarrow$ \\
\hline
Qwen3-32B (step~144) & 0.1979 & 11.6\% & 0.0287 \\
GPT-5 & 0.1986 & 11.3\% & 0.0812 \\
Naive Baseline & 0.2238 & 0\% & --- \\
Qwen3-32B (base) & 0.2381 & $-6.4$\% & 0.1419 \\
\hline
\end{tabular}
\caption{Aggregate probabilistic performance on the held-out test set. Lower Brier score and ECE indicate better performance, while higher Brier skill score is better.}
\label{tab:aggregate-performance}
\end{table}

The fine-tuned model outperforms both the pretrained base model, the naive baseline, and a frontier general-purpose model across all metrics. Fine-tuning yields improvements in both probabilistic accuracy and calibration. Relative to the pretrained base model, the reduction in ECE represents a nearly 80\% reduction in calibration error.

\begin{figure}[htbp]         
  \centering
  \includegraphics[width=\linewidth]{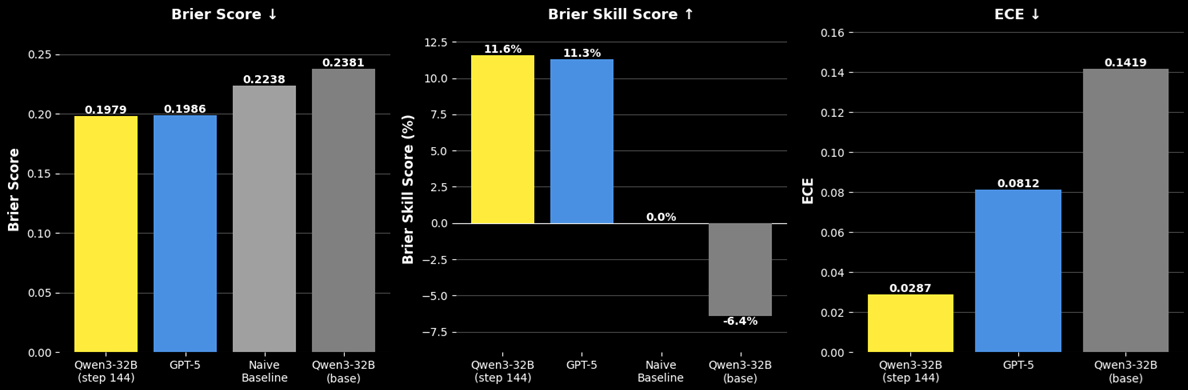}
  \caption{Aggregate performance on the held-out test set}
  \label{fig:aggregate-performance}
\end{figure}

\subsection{Calibration Behavior}

Calibration improvements are a central outcome of training with Foresight Learning. Expected calibration error decreases from 0.1419 for the pretrained base model to 0.0287 after fine-tuning, indicating substantially better alignment between predicted probabilities and realized outcomes. In contrast, the naive baseline exhibits no discriminative power and cannot be meaningfully calibrated beyond the base rate.

\begin{figure}[htbp]         
  \centering
  \includegraphics[width=0.5\linewidth]{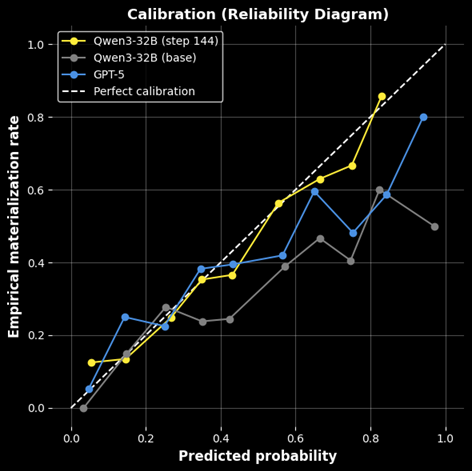}
  \caption{Reliability diagram on the test set showing empirical materialization rates as a function of predicted risk probabilities.}
  \label{fig:calibration}
\end{figure}

\ref{fig:calibration} presents a reliability diagram on the test set. Predicted probabilities from the trained model closely track empirical materialization rates across the score range, and higher predicted risk probabilities correspond to higher observed materialization frequencies. This monotonic relationship indicates that the model produces well-ordered and interpretable probability estimates rather than merely binary classifications.

\subsection{Performance by Risk Category}

Table~\ref{tab:category-performance} reports Brier scores by high-level risk category for the trained model (step~144) and the pretrained base model. Lower values indicate better probabilistic calibration. Categories are ordered by sample count.

\begin{table}[t]
\centering
\begin{tabular}{lccc}
\hline
Risk Category & Step~144 & Base & Count \\
\hline
Expenses, Costs \& Balance Sheet Metrics & 0.2120 & 0.2525 & 104 \\
Governance, Compliance \& Internal Controls & 0.1615 & 0.2058 & 81 \\
Revenue \& Sales Performance & 0.2527 & 0.2481 & 68 \\
Strategic Actions \& Corporate Transactions & 0.2166 & 0.2977 & 68 \\
Profitability \& Margins & 0.2342 & 0.2659 & 55 \\
Legal \& Regulatory Events & 0.1091 & 0.1742 & 41 \\
Other & 0.1466 & 0.1802 & 29 \\
Regulatory Approvals \& Product Development & 0.2138 & 0.2328 & 28 \\
Operational Events \& Disruptions & 0.1355 & 0.2118 & 26 \\
\hline
\end{tabular}
\caption{Brier scores by high-level risk category for the trained model (step~144) and the pretrained base model. Lower values indicate better probabilistic calibration. Categories are ordered by sample count.}
\label{tab:category-performance}
\end{table}

Across categories, the trained model achieves lower Brier scores than the base model in 8 of 9 risk types, indicating improved calibration across most classes of risks. The largest improvements occur for Strategic Actions \& Corporate Transactions, Operational Events \& Disruptions, and Legal \& Regulatory Events, where outcomes tend to correspond to discrete, explicitly disclosed events.

In absolute terms, the lowest Brier scores are observed for Legal \& Regulatory Events and Operational Events \& Disruptions, while financially oriented categories such as Revenue \& Sales Performance and Profitability \& Margins exhibit higher Brier scores. This pattern suggests that risks tied to clearly identifiable events are easier to probabilistically resolve under the proposed labeling and time-horizon framework, whereas financial outcomes often depend on degree, attribution, or cumulative effects that are less sharply disclosed. Because category sizes vary substantially, particularly for lower-frequency categories, these per-category estimates should be interpreted accordingly.

\section{Discussion}

Our results demonstrate that large language models trained with Foresight Learning can extract meaningful, forward-looking signal from unstructured SEC risk disclosures and express that signal as well-calibrated probabilities. These gains do not rely on proprietary data, manual annotation, or domain-specific feature engineering. Instead, they arise from a fully automated data generation pipeline that operates entirely on public SEC filings, generating firm-specific, time-bounded risk queries and resolving their outcomes against subsequent disclosures. This process produces large-scale, risk-level supervision that would be infeasible to construct manually, enabling calibrated probabilistic learning directly from narrative regulatory text.

The synthetic data generation pipeline introduced in this work provides a general mechanism for transforming unstructured chronological disclosures into structured, outcome-linked training data, enabling probabilistic learning at a granularity that would be infeasible to label manually. Training on more than 6,000 such automatically generated samples is sufficient to induce meaningful probabilistic behavior in a compact model.
Notably, the trained model outperforms a frontier large language model on both accuracy and calibration, despite being orders of magnitude smaller. This result suggests that frontier scale alone is not sufficient for probabilistic reasoning in specialized domains, and that task-aligned supervision can outweigh sheer model capacity.

This finding has direct implications for deployment in financial and regulatory settings. Many institutions operate under constraints that prevent sharing sensitive data or internal strategies with external model providers, limiting the applicability of closed frontier models. Our results demonstrate that compact, domain-specific models trained exclusively on chronological documents can exceed frontier performance while remaining small enough to run on a single GPU, enabling practical use in data-sensitive environments.

Performance varies across risk categories. Risks tied to discrete, explicitly disclosed events - such as legal actions, regulatory developments, and operational disruptions - are predicted more accurately and with better calibration than diffuse financial outcomes. This pattern reflects differences in how clearly outcomes are disclosed rather than limitations of the modeling approach.

Overall, this work presents a scalable approach for training expert domain models that outperform frontier systems while requiring no manual labeling, no proprietary data, and orders-of-magnitude fewer parameters. More broadly, it suggests that access to temporally grounded supervision, rather than model scale alone, is a key driver of improved world modeling.

\section{Limitations and Future Work}

This work has several limitations. First, materialization labels are derived automatically from future SEC filings and are therefore disclosure-dependent and noisy. A risk may materialize without being explicitly disclosed, may be disclosed ambiguously, or may be disclosed in a way that is difficult to link unambiguously to the original risk query. As a result, the task predicts the probability that a risk is disclosed as having materialized, rather than the true occurrence or severity of the underlying event, and predictions should be interpreted as conditional on firms’ reporting behavior and disclosure incentives.

Second, the dataset focuses on relatively short prediction horizons and summary-level risk context, which may under-represent long-term, slowly evolving, or highly diffuse risks. Firms may also strategically revise risk disclosures over time, potentially introducing systematic biases that the model learns.

Many of these limitations stem from the structure and scope of available disclosures rather than from the modeling approach itself. Extending the data generation pipeline to longer horizons, richer contextual inputs, and improved outcome linking offers a natural path to addressing them. More broadly, this work points toward a general paradigm for learning from unstructured, time-ordered text in settings where labels are scarce or subjective: using the future itself as supervision. Applying this methodology to other disclosure regimes and narrative reporting domains may enable models to learn not only calibrated probabilities, but richer forms of temporal reasoning, abstraction, and decision-relevant understanding directly from natural language.

\bibliographystyle{tmlr}  
\bibliography{Ref}   

@article{turtel2026,
  title={Future-as-Label: Scalable Supervision from Real-World Outcomes},
  author={Turtel, Benjamin and Wilczewski, Paul and Franklin, Danny and Skothiem, Kris},
  journal={arXiv preprint arXiv:2601.06336},
  year={2026},
  url={https://arxiv.org/abs/2601.06336}
}

@article{caillaut2025,
  title={The LLM Pro Finance Suite: Multilingual Large Language Models for Financial Applications},
  author={Caillaut, Ga{\"e}tan and Qader, Raheel and Liu, Jingshu and Nakhl{\'e}, Mariam and Sadoune, Arezki and Ahmim, Massinissa and Barthelemy, Jean-Gabriel},
  journal={arXiv preprint arXiv:2511.08621},
  year={2025},
  url={https://arxiv.org/pdf/2511.08621}
}

@article{kesari2022,
  title={Predicting Cybersecurity Incidents Through Mandatory Disclosure Regulation},
  author={Kesari, Aniket},
  journal={Illinois Journal of Law, Technology \& Policy},
  year={2020},
  url={https://www.law.berkeley.edu/wp-content/uploads/2020/10/Predicting_Cybersecurity_Incidents_Through_Mandatory_Disclosure_Regulations.pdf}
}

@article{zavitsanos2025,
  title={Machine Learning for Identifying Risk in Financial Statements: A Survey},
  author={Zavitsanos, Elias and Spyropoulou, Eirini and Giannakopoulos, George},
  journal={ACM Computing Surveys},
  volume={57},
  number={9},
  year={2025},
  url={https://dl.acm.org/doi/10.1145/3723157}
}

@article{hajekmunk2025,
  title={Corporate financial distress prediction using the risk-related information content of annual reports}, 
  author={Hajek, Petr and Munk, Michal}, 
  journal={Information Processing \& Management},
  volume={61},
  number={5},
  year={2025},
  url={https://www.sciencedirect.com/science/article/abs/pii/S0306457324001791}
}

\end{document}